\documentclass[runningheads]{llncs}
\usepackage[T1]{fontenc}
\usepackage{lmodern}
\usepackage{graphicx}
\usepackage{amsmath}
\usepackage[colorlinks=true,linkcolor=blue!60!black,citecolor=blue!60!black,urlcolor=blue!60!black]{hyperref}
\usepackage{orcidlink}

\graphicspath{{figures/}}
\begin{document}

\title{How Far Can Sub-3B Open Language Models Go in Zero-Shot Essay
Scoring on an 8\,GB Consumer GPU?}

\titlerunning{Sub-3B Zero-Shot Essay Scoring on a Consumer GPU}
\author {Nguyen Dung Son\inst{1}\orcidlink{0009-0005-6315-9339} \and
Dang Quang Minh\inst{1}\orcidlink{0009-0000-9113-6680} \and
Nguyen Huu Loi\inst{2}\orcidlink{0000-0001-7987-0348}\and
Truong Viet Vu\inst{3}\orcidlink{0009-0004-3941-1306} \and
Nguyen Thai Anh\inst{3}\orcidlink{0009-0005-5600-6510}\thanks{Corresponding author.}}
\institute {FPT Polyschool, FPT University, Hanoi, Vietnam \and
FPT Polyschool, FPT University, Can Tho, Vietnam\and
Faculty of Information Technology, Van Lang School of Technology,\\
Van Lang University, Ho Chi Minh City, Vietnam\\
\email{anh.nt@vlu.edu.vn}}
\authorrunning{N. D. Son et al.}

\maketitle

\begin{abstract}
Zero-shot essay scoring with large language models is usually demonstrated
with proprietary API models, yet the settings where automated scoring is
most needed, such as public schools grading thousands of essays under strict
privacy rules, are often the settings where sending student writing to a
third-party API is not acceptable. We ask how much of that capability
survives when the model must be a sub-3B open model running fully locally
in FP16, and answer
with a controlled study of four instruction-tuned models from two families
(Qwen2.5 at 0.5B/1.5B/3B and SmolLM2 at 1.7B) on all eight ASAP-AES prompts,
scored entirely on a single 8\,GB consumer GPU, with bootstrap confidence
intervals, Holm-corrected paired tests, and deployment-realistic variants of
the key design choices. Three findings emerge. \emph{(i)} Rubric-decomposed
prompting beats holistic prompting for every model under batch min-max
aggregation (though not on every prompt: Qwen2.5-3B decreases
significantly on one), and under mean aggregation two unrelated
families land within 0.01 of each other at the 1.5--1.7B scale. \emph{(ii)}
The mapping of trait scores into the prompt score range is fragile to grader
calibration: one model compresses trait scores into a narrow low band
(2--4 on the 0--10 scale) and naive mean aggregation collapses, while the min-max normalization
of Multi-Trait Specialization repairs it (macro QWK 0.204 to 0.388) and
remains within 0.03 when its statistics are frozen on 30 held-out essays.
\emph{(iii)} Signed error falls with essay length in eleven of twelve
configurations; under holistic prompting the two smaller Qwen models cross
from overrating to underrating, SmolLM2 declines to neutral, and
Qwen2.5-3B underrates at every length, opposite to the verbosity bias
reported for large LLM judges; rubric decomposition with normalization
largely flattens this slope for the well-calibrated models. We anchor all results honestly: the best local configuration
(0.388) remains far below both the human inter-rater ceiling (0.769) and
even a length-only baseline (0.523, itself normalized with the same batch
min-max), so we position sub-3B local models strictly for formative,
human-supervised feedback.
\end{abstract}

\section{Introduction}\label{sec:intro}

Automated essay scoring (AES) is one of the oldest ambitions of computers in
education, dating back to Page's Project Essay Grade in the 1960s
\cite{page1966}, and for most of its history it has required supervised
training: a model is fitted to hundreds or thousands of human-scored essays
for one specific prompt \cite{taghipour2016,dong2017}. Large language models
(LLMs) changed the economics of the task. A sufficiently capable
instruction-tuned model can grade an essay it has never seen, guided only by
a description of the writing task and the rubric
\cite{mizumoto2023,lee2024mts}. The strongest zero-shot and few-shot results
to date are reported with proprietary API models \cite{yancey2023}, although
trait decomposition already lets open models in the 7B--13B range
(e.g.\ Mistral-7B, Llama2-13B) rival ChatGPT on this task \cite{lee2024mts}.

There is, however, a structural mismatch between where this capability lives
and where it is needed. The classrooms that would benefit most from
automated formative feedback, large public-school classes and institutions
in developing regions, are often precisely the settings where transmitting
student writing to a third-party service conflicts with privacy regulation
or school policy; insufficient privacy consideration is among the most
consistently flagged concerns in the literature on LLMs in education
\cite{yan2024}. Efficiency is a scientific concern as well: results that
only exist at the largest scale are hard to reproduce, audit, or deploy
\cite{schwartz2020}. A capability that exists only behind an API does not
transfer to these deployments, and the literature offers little guidance on
what remains of zero-shot AES when the model must be small, open, and local.

\begin{figure}[tbp]
\centering
\includegraphics[width=\textwidth]{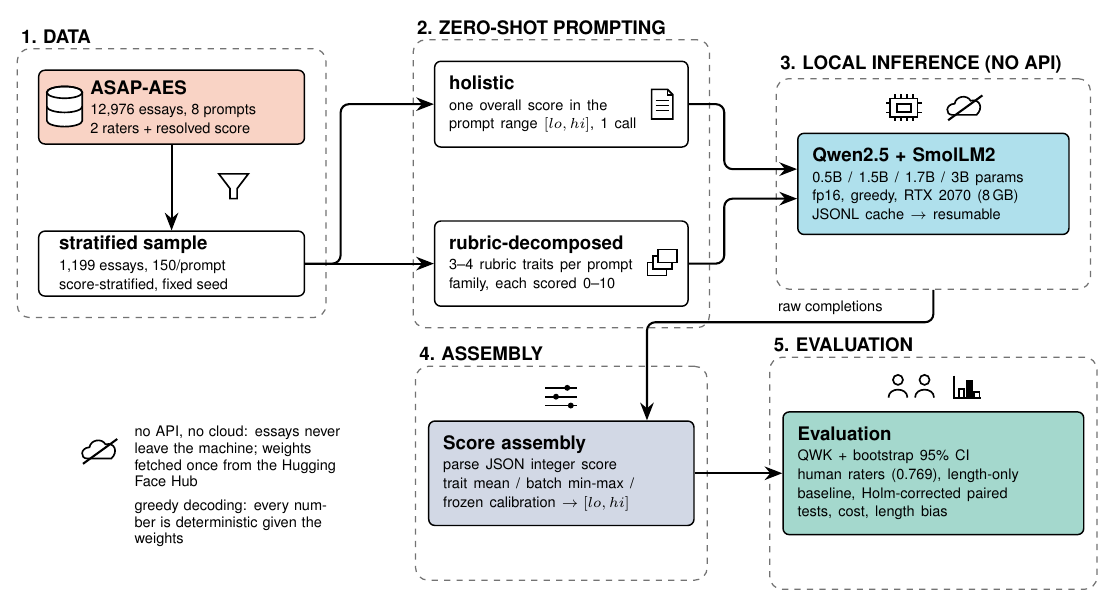}
\caption{\textbf{Study overview.} A score-stratified sample of ASAP-AES
essays (1{,}199 essays, all eight prompts) is scored under two prompting
strategies by four small open models hosted entirely on one 8\,GB consumer
GPU; trait scores are assembled into prompt-range scores and evaluated
against the resolved human score with bootstrap confidence intervals, paired
tests, cost accounting, and length-bias analysis. Every model call is
cached, so interrupted runs resume without recomputation.}
\label{fig:pipeline}
\end{figure}

This paper addresses that gap for sub-3B FP16 open
models\footnote{We write sub-3B for open models of at most 3B nominal
parameters; the largest we test is Qwen2.5-3B.} with a
systematic study under the strictest deployment constraint we can
operationalize (Fig.~\ref{fig:pipeline}):
every token is generated on one consumer GPU with 8\,GB of memory, no API is
called at any point, and model weights are fetched once from a public hub.
Within this budget we compare holistic prompting, which asks for a single
score in the official range of each prompt, against rubric-decomposed
prompting in the style of Multi-Trait Specialization (MTS)
\cite{lee2024mts}, which scores rubric traits in separate calls and
aggregates them. The study covers four instruction-tuned models from two
independent families and all eight ASAP-AES prompts, and reports uncertainty
for every headline comparison: bootstrap confidence intervals up to the
macro level, paired tests with Holm correction and directional counts, the
agreement of the two human raters on the same essays as the reference
ceiling, and a length-only baseline as the skeptic's floor.

Beyond confirming that rubric decomposition, under batch min-max
aggregation, transfers to the sub-3B regime,
the study surfaces two effects that become acute at this scale. The
first is \emph{grader harshness}: one model compresses its trait scores
into a narrow low band (2--4 on the 0--10 scale), and the standard mapping of trait means into
the prompt score range silently collapses; the min-max normalization that
MTS introduced for exactly this reason repairs it, but is not universally
superior, so aggregation must be treated as a calibration-dependent choice.
The second is a systematic \emph{length effect}: signed error declines with
essay length in nearly every configuration; under holistic prompting the
two smaller Qwen models cross from overrating to underrating, SmolLM2
declines to neutral, and Qwen2.5-3B underrates at every length, a drift
opposite to the verbosity bias documented for large LLM judges
\cite{zheng2023}; rubric decomposition with normalization largely flattens
the slope for the well-calibrated models. Both effects have direct consequences for anyone
deploying local LLM scoring at this scale, and both would be easy to miss
without per-trait and per-quartile analysis.

Our contributions are:
\begin{enumerate}
  \item The first systematic zero-shot AES benchmark, to our knowledge, of
  sub-3B open models under a fully local deployment constraint, covering
  Qwen2.5 (0.5B, 1.5B, 3B) and SmolLM2 (1.7B) on all eight ASAP-AES prompts
  with 21{,}584 scored model calls and one parse failure.
  \item Cross-family evidence that rubric decomposition is the most
  effective lever among the strategies we test: under batch min-max
  aggregation it improves macro QWK for every model, with Holm-corrected
  significantly positive differences on up to six of eight prompts (and
  one significantly negative prompt for Qwen2.5-3B), and two
  unrelated families converge to nearly the same trait-prompted accuracy
  (within 0.01 under mean aggregation) at the 1.5--1.7B scale.
  \item A characterization of how trait aggregation interacts with grader
  calibration. The min-max normalization itself is due to MTS; what we document is why it matters at small scale
  (grader harshness makes naive mean aggregation non-monotonic in model
  size), that it is not universally superior, and that a deployment-safe
  frozen-calibration variant retains most of its benefit (here and
  throughout, this calibration is score-range alignment, not probability
  calibration in the traditional sense).
  \item An honest accounting of where this capability stands: a length-bias
  analysis with the opposite sign to LLM-judge verbosity bias, a
  length-only baseline (0.523, computed with the same transductive batch
  min-max) that still exceeds every LLM configuration, per-essay cost in
  calls, tokens, and seconds, and mean absolute errors in raw score
  points.
\end{enumerate}

\section{Related Work}\label{sec:related}

\textbf{Automated essay scoring.} AES predates modern NLP by decades
\cite{page1966} and has been deployed operationally since e-rater
\cite{attali2006}. The field standardized around supervised learning on
human-scored corpora: discriminative ranking on ESOL texts
\cite{yannakoudakis2011}, and, after the release of the ASAP-AES benchmark
\cite{asap2012}, a progression of neural architectures from LSTM- and
CNN-based regressors \cite{alikaniotis2016,taghipour2016} to
attention-based models \cite{dong2017} that approach human-level agreement
on several prompts. Comprehensive surveys trace this development
\cite{ke2019,uto2021,ramesh2022}. A recurring critique of the field is the
length confound: essay length alone predicts human scores remarkably well,
which has been used to question what AES systems actually measure
\cite{perelman2014}. We take this critique seriously and report a
length-only baseline.

\textbf{Zero-shot and few-shot LLM scoring.} Prompting removes the training
requirement. Mizumoto and Eguchi score 12{,}100 TOEFL essays with GPT-3 and
find useful but clearly sub-human reliability \cite{mizumoto2023}; Yancey
et al.\ show that GPT-4 with calibration examples approaches the agreement
of modern automated writing evaluation systems on L2 essays, while noting
that agreement varies with the writer's first language \cite{yancey2023}.
Prompting strategies matter: Stahl et al.\ study joint scoring and feedback
prompting \cite{stahl2024}, benchmarks now probe multimodal essay scoring
\cite{essayjudge2025}, and, most relevant here, Lee et al.\ show on ASAP
that multi-trait decomposition with min-max scaling substantially narrows
the zero-shot gap for ChatGPT and for 7B--13B open models
\cite{lee2024mts}. Separately, the LLM-as-judge literature documents a
verbosity bias in which longer responses are systematically favored
\cite{zheng2023}. Our study differs from prior AES prompting work in three
ways: the models are up to an order of magnitude smaller (0.5B--3B) and run
fully locally on an 8\,GB GPU; the comparison is replicated across two
unrelated model families; and the headline differences carry bootstrap
confidence intervals and Holm-corrected paired tests, which prior prompting
studies rarely provide.

\textbf{Small open models and efficiency.} The Green AI position paper
argued that results obtainable only at the largest scale carry a scientific
and equity cost \cite{schwartz2020}. Since then, sub-3B instruction-tuned
models trained on trillion-token curricula, such as the Qwen2.5 series
\cite{qwen25} and SmolLM2 \cite{smollm2}, have become reliable enough to
follow structured output formats consistently; in our study they produce
exactly one unparseable response in 21{,}584 calls. This reliability, more
than raw capability, is what makes an all-local AES pipeline practical.

\textbf{Positioning.} To our knowledge no prior work quantifies zero-shot
AES for sub-3B local models across two families with corrected significance
testing, characterizes the interaction between trait aggregation and grader
calibration including a deployment-safe variant, or measures how prompting
strategy shapes length bias at this scale. These are the gaps this paper
fills.

\section{Method}\label{sec:method}

\subsection{Prompting Strategies}
\textbf{Holistic.} The model receives a one-sentence description of the
writing task, the essay, and the official score range $[lo, hi]$ of the
prompt, and must return one integer score as a JSON object of the form
\texttt{\{"score": <int>\}}. We request JSON rather than free text because
it makes parsing deterministic and auditable; a regex fallback recovers
scores from near-JSON outputs, anything else counts as a parse failure
attributed to the method, and parsed values outside the requested range are
clamped to it (this occurred in 2 of 21{,}584 calls).

\textbf{Rubric-decomposed (traits).} The model scores one trait per call on
a 0--10 integer scale, with an instruction to ignore all other aspects of
the essay. Trait sets follow the ASAP rubric families: ideas and content,
organization, style and word choice, and conventions for the persuasive
prompts (1, 2); ideas and content, organization, voice and word choice, and
conventions for the narrative prompts (7, 8); content understanding, use of
the source text, and clarity for the source-dependent prompts (3--6). The
full templates are given in Appendix~A. We use a fixed 0--10 trait scale
rather than each prompt's native range so that trait behavior is comparable
across prompts and models; Sect.~\ref{sec:results} shows this choice is
exactly where calibration effects surface.

\subsection{Score Assembly}\label{sec:assembly}
Let $t_1, \dots, t_k$ be the trait scores of an essay and
$m = \frac{1}{k}\sum_i t_i \in [0, 10]$ its trait mean. The \emph{mean}
aggregation maps $m/10$ linearly into $[lo, hi]$. The \emph{batch min-max}
aggregation rescales $m$ across the evaluated essays of the same prompt to
span $[0, 1]$ before the same mapping; this follows MTS \cite{lee2024mts}
and makes the final score invariant to a model's global harshness or
leniency on the trait scale. Because the trait-mean lattice mapped onto
coarse score ranges produces many exact half-way ties, we fix the rounding
convention explicitly (ties round half up) and apply it identically in
every analysis. Batch min-max is transductive; we therefore also evaluate a
\emph{frozen-calibration} variant in which the normalization statistics are
estimated once on 30 held-out essays per prompt and applied unchanged to
the rest (Sect.~\ref{sec:results}); we report the mean over 200 random
choices of the held-out set (Table~\ref{tab:robust}). All aggregations reuse identical cached
model calls, so their comparison is exact, not resampled.

\subsection{Models and Local Inference}
We evaluate four instruction-tuned models spanning 0.5B to 3B parameters
and two independent families: Qwen2.5-Instruct at 0.5B, 1.5B, and 3B
\cite{qwen25}, and SmolLM2-1.7B-Instruct \cite{smollm2}. The two families
differ in pretraining corpus, tokenizer, and instruction tuning, which lets
us separate effects of the prompting strategy from idiosyncrasies of one
model line. All models run in FP16 with greedy decoding, so every model
output is deterministic given the weights, and all resampling analyses use
fixed seeds. Hardware is one NVIDIA RTX 2070 Max-Q with 8\,GB of memory. Weights are fetched once from
the Hugging Face Hub; afterwards no network access is needed and essays
never leave the machine. Every call is appended to a JSONL cache keyed by
model, method, essay, and trait, so an interrupted run resumes exactly
where it stopped; the full study is reproducible from this cache without
any model call.

\section{Experimental Setup}\label{sec:setup}
\textbf{Data.} We use ASAP-AES \cite{asap2012}, the standard AES benchmark:
12{,}976 essays in the public training release, over eight prompts spanning
persuasive, source-dependent, and narrative writing, each with two
independent human rater scores and a resolved score. Score ranges differ
per prompt (from 0--3 to 0--60), which is why all metrics are computed per
prompt. We evaluate on a score-stratified sample of 150 essays per prompt
(1{,}199 total, fixed seed; prompt 4 contributes 149), so the sample
preserves each prompt's score distribution; the per-call cache allows
scaling to the full corpus.

\textbf{Metrics.} Agreement is measured by quadratic weighted kappa
\cite{cohen1968}, computed per prompt over its official score range; macro
QWK is the unweighted mean over the eight prompts. The human reference is
the rater~1 versus rater~2 QWK on the same sampled essays, computed on the
raters' native score scale. We compute essay-level bootstrap 95\% confidence
intervals for per-prompt QWK, a
macro-level bootstrap (essays resampled
within every prompt, all eight QWKs and their mean recomputed, with the
min-max normalization re-estimated inside every resample), and paired
method comparisons on identical resamples with Holm correction within each
family of eight prompts and directional significance counts. Cost is
accounted per essay (calls, prompt and completion tokens, wall-clock
seconds), length bias as the mean signed error normalized by the prompt
score range within essay-length quartiles, and we additionally report mean
absolute error in raw score points.

\section{Results}\label{sec:results}

\textbf{Rubric decomposition helps every model under batch min-max
aggregation, with one calibration caveat.} Table~\ref{tab:macro} gives the
headline comparison and Table~\ref{tab:delta} the corrected paired tests.
Under batch min-max aggregation, trait prompting improves macro QWK over
holistic prompting for every model: $+0.137$ for Qwen2.5-0.5B (Holm-corrected significantly
positive on 1 of 8 prompts), $+0.192$ for Qwen2.5-1.5B (5/8), $+0.103$ for
SmolLM2-1.7B (2/8), and $+0.170$ for Qwen2.5-3B (6/8, with 1 significantly
negative). Under the naive mean the gain holds for three of four models but
reverses for Qwen2.5-3B ($-0.013$), for reasons the calibration analysis
below makes precise. At the 1.5--1.7B scale the two families land within 0.01 of
each other under mean aggregation (0.336 versus 0.345) despite entirely
different pretraining corpora and instruction tuning \cite{qwen25,smollm2},
which suggests the effect is a property of the prompting strategy rather
than of one model family. The mechanism is intuitive: a small model asked
for one holistic number must implicitly weigh many rubric dimensions at
once, while trait prompting decomposes the judgment into narrower questions
that a small model can answer more reliably.

\begin{table}[tbp]
\centering
\caption{Macro QWK over the eight ASAP prompts (1{,}199 essays), mean parse-failure rate, and wall-clock cost per essay on a single RTX~2070 Max-Q (8\,GB). Trait scores use the same cached model calls for both aggregations.}
\label{tab:macro}
\begin{tabular}{llrrr}
\hline
Model & Prompting & QWK & Fail rate & s/essay \\
\hline
Qwen2.5-0.5B & traits (mean) & 0.147 & 0.001 & 1.58 \\
Qwen2.5-0.5B & traits (min-max) & 0.195 & 0.001 & 1.58 \\
Qwen2.5-0.5B & holistic & 0.056 & 0.000 & 0.46 \\
Qwen2.5-1.5B & traits (mean) & 0.336 & 0.000 & 3.77 \\
Qwen2.5-1.5B & traits (min-max) & 0.353 & 0.000 & 3.77 \\
Qwen2.5-1.5B & holistic & 0.162 & 0.000 & 0.59 \\
SmolLM2-1.7B & traits (mean) & 0.345 & 0.000 & 2.90 \\
SmolLM2-1.7B & traits (min-max) & 0.331 & 0.000 & 2.90 \\
SmolLM2-1.7B & holistic & 0.229 & 0.000 & 0.82 \\
Qwen2.5-3B & traits (mean) & 0.204 & 0.000 & 2.85 \\
Qwen2.5-3B & traits (min-max) & 0.388 & 0.000 & 2.85 \\
Qwen2.5-3B & holistic & 0.217 & 0.000 & 0.80 \\
\hline
Human raters & rater 1 vs rater 2 & 0.769 & -- & -- \\
\hline
\end{tabular}
\end{table}

\begin{table}[tbp]
\centering
\caption{Paired bootstrap comparison against holistic prompting (identical resamples, 2{,}000 iterations, min-max normalization re-estimated inside every resample): mean QWK difference across prompts and the number of prompts (out of 8) with a significantly positive or negative difference after Holm correction.}
\label{tab:delta}
\begin{tabular}{llrrr}
\hline
Model & Comparison & $\Delta$QWK & Sig.$+$ & Sig.$-$ \\
\hline
Qwen2.5-0.5B & traits (mean) vs holistic & $+0.088$ & 2/8 & 0/8 \\
Qwen2.5-0.5B & traits (min-max) vs holistic & $+0.137$ & 1/8 & 0/8 \\
Qwen2.5-1.5B & traits (mean) vs holistic & $+0.174$ & 5/8 & 0/8 \\
Qwen2.5-1.5B & traits (min-max) vs holistic & $+0.192$ & 5/8 & 0/8 \\
SmolLM2-1.7B & traits (mean) vs holistic & $+0.116$ & 3/8 & 0/8 \\
SmolLM2-1.7B & traits (min-max) vs holistic & $+0.103$ & 2/8 & 0/8 \\
Qwen2.5-3B & traits (mean) vs holistic & $-0.013$ & 3/8 & 2/8 \\
Qwen2.5-3B & traits (min-max) vs holistic & $+0.170$ & 6/8 & 1/8 \\
\hline
\end{tabular}
\end{table}

\begin{figure}[tbp]
\centering
\includegraphics[width=\textwidth]{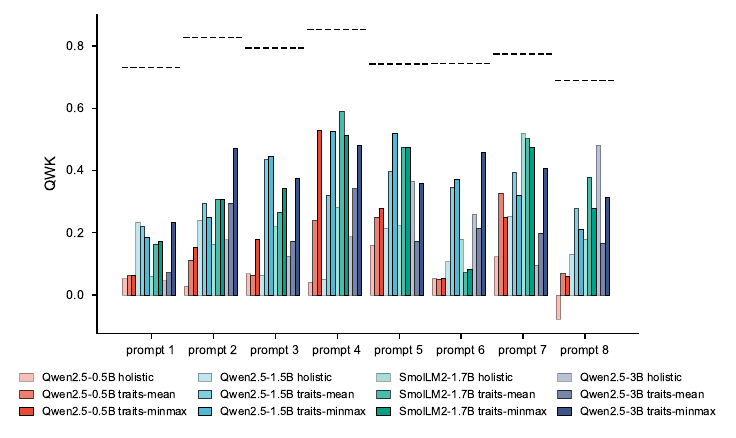}
\caption{\textbf{QWK per ASAP prompt} for all models and prompting
strategies (bar opacity encodes the aggregation); dashed segments mark the
human rater agreement per prompt.}
\label{fig:perset}
\end{figure}

\textbf{Per-prompt profile.} Table~\ref{tab:perset} breaks the comparison
down by prompt, visualized in Fig.~\ref{fig:perset}. The source-dependent
prompt 4 is the easiest for all four models (QWK 0.48 to 0.53 with min-max
trait prompting, against a human ceiling of 0.85), and the source-dependent
group benefits most consistently from trait decomposition, plausibly
because their rubrics reduce to verifiable questions about the source text.
The hardest prompt depends on the model: the persuasive prompt 1 for the
two larger Qwen models, prompt 6 for Qwen2.5-0.5B and SmolLM2-1.7B (QWK
below 0.09). Per-prompt differences of this size sit inside overlapping
per-prompt confidence intervals, so we treat them as descriptive only.

\begin{table}[tbp]
\centering
\caption{QWK per ASAP prompt (P1--P8); holistic prompting versus rubric-decomposed prompting with batch min-max aggregation. The last row is the agreement of the two human raters on the same essays.}
\label{tab:perset}
\setlength{\tabcolsep}{3.4pt}
\begin{tabular}{lrrrrrrrrr}
\hline
Model & P1 & P2 & P3 & P4 & P5 & P6 & P7 & P8 & Macro \\
\hline
\multicolumn{10}{l}{\emph{holistic}} \\
Qwen2.5-0.5B & 0.05 & 0.03 & 0.07 & 0.04 & 0.16 & 0.05 & 0.12 & -0.08 & 0.056 \\
Qwen2.5-1.5B & 0.23 & 0.24 & 0.06 & 0.05 & 0.21 & 0.11 & 0.25 & 0.13 & 0.162 \\
SmolLM2-1.7B & 0.06 & 0.16 & 0.22 & 0.28 & 0.22 & 0.18 & 0.52 & 0.18 & 0.229 \\
Qwen2.5-3B & 0.05 & 0.18 & 0.13 & 0.19 & 0.37 & 0.26 & 0.10 & 0.48 & 0.217 \\
\hline
\multicolumn{10}{l}{\emph{traits, batch min-max}} \\
Qwen2.5-0.5B & 0.06 & 0.15 & 0.18 & 0.53 & 0.28 & 0.05 & 0.25 & 0.06 & 0.195 \\
Qwen2.5-1.5B & 0.19 & 0.25 & 0.45 & 0.53 & 0.52 & 0.37 & 0.32 & 0.21 & 0.353 \\
SmolLM2-1.7B & 0.17 & 0.31 & 0.34 & 0.51 & 0.48 & 0.08 & 0.48 & 0.28 & 0.331 \\
Qwen2.5-3B & 0.23 & 0.47 & 0.37 & 0.48 & 0.36 & 0.46 & 0.41 & 0.31 & 0.388 \\
\hline
Human raters & 0.73 & 0.83 & 0.79 & 0.85 & 0.74 & 0.74 & 0.77 & 0.69 & 0.769 \\
\hline
\end{tabular}
\end{table}

\begin{figure}[tbp]
\centering
\includegraphics[width=\textwidth]{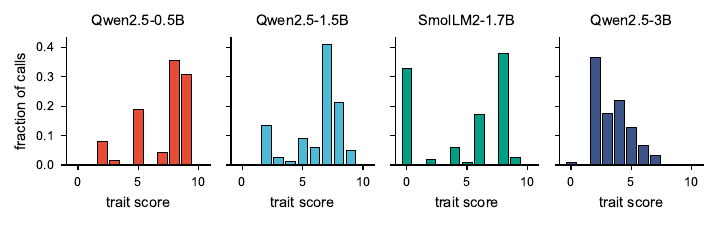}
\caption{\textbf{Grader calibration.} Distribution of raw trait scores
(0--10 scale, all prompts and traits pooled) per model. The two smaller
Qwen models are lenient (modes at 7--9), SmolLM2 spreads widely (mean 4.7,
with a third of calls at 0), and Qwen2.5-3B concentrates its mass on 2--4
(mean 3.4); naive mean aggregation inherits these offsets, batch min-max
removes them.}
\label{fig:traitdist}
\end{figure}

\textbf{Aggregation interacts with grader calibration.}
Fig.~\ref{fig:traitdist} shows the mechanism behind our second finding: the
four models use the same 0--10 trait scale in radically different ways.
Qwen2.5-3B assigns trait scores almost exclusively in the 2--4 band (mean
3.4, s.d.\ 1.5), so naive mean aggregation compresses its predictions into
the bottom of each prompt range and its macro QWK collapses to 0.204,
erasing the benefit of its extra parameters. Min-max normalization, the
device MTS introduced for exactly this failure mode \cite{lee2024mts},
restores 0.388, the best local result in the study, and yields a consistent
ordering with scale (0.195, 0.353, 0.388 for 0.5B, 1.5B, 3B), although the
macro-level bootstrap intervals of the top two configurations overlap
(Table~\ref{tab:robust}), so we do not claim the 3B advantage over 1.5B is
significant. The ablation in Table~\ref{tab:agg} shows this ordering holds
only under batch min-max: no alternative aggregation rescues Qwen2.5-3B
(median 0.227, drop-min 0.247), which is the non-monotonicity noted in
contribution 3. The ablation also shows the repair is not a free
lunch: for SmolLM2-1.7B, whose wide trait distribution needs no repair, the
plain mean is slightly better (0.345 versus 0.331). Aggregation should
therefore be treated as a calibration-dependent design choice, selected
after inspecting the trait score distribution on a handful of essays.

\begin{table}[tbp]
\centering
\caption{Deployment robustness of trait scoring with min-max aggregation: transductive (normalized on the evaluated batch, with macro-level bootstrap 95\% CI) versus frozen calibration (normalization estimated on 30 held-out essays per prompt, mean of 200 random splits). The length-only baseline maps essay length through the same batch min-max onto each prompt range.}
\label{tab:robust}
\begin{tabular*}{\textwidth}{@{\extracolsep{\fill}}lccc}
\hline
Model & Transductive & Macro 95\% CI & Frozen (30) \\
\hline
Qwen2.5-0.5B & 0.195 & [0.158, 0.234] & 0.206 \\
Qwen2.5-1.5B & 0.353 & [0.312, 0.404] & 0.362 \\
SmolLM2-1.7B & 0.331 & [0.287, 0.366] & 0.319 \\
Qwen2.5-3B & 0.388 & [0.332, 0.421] & 0.361 \\
\hline
length-only baseline & 0.523 & -- & -- \\
\hline
\end{tabular*}
\end{table}

\begin{table}[tbp]
\centering
\caption{Trait-aggregation ablation (macro QWK). All variants reuse identical cached trait scores; only the mapping into the prompt range changes.}
\label{tab:agg}
\begin{tabular*}{\textwidth}{@{\extracolsep{\fill}}lrrrr}
\hline
Model & mean & median & drop-min & batch min-max \\
\hline
Qwen2.5-0.5B & 0.147 & 0.135 & 0.131 & 0.195 \\
Qwen2.5-1.5B & 0.336 & 0.332 & 0.316 & 0.353 \\
SmolLM2-1.7B & 0.345 & 0.340 & 0.339 & 0.331 \\
Qwen2.5-3B & 0.204 & 0.227 & 0.247 & 0.388 \\
\hline
\end{tabular*}
\end{table}

\textbf{Deployment robustness and the length-only floor.}
Table~\ref{tab:robust} answers the two questions a skeptical deployer
should ask. First, does the min-max benefit survive without transductive
access to the evaluation batch? Largely yes: freezing the normalization on
30 held-out essays per prompt costs Qwen2.5-3B 0.027 (0.388 to 0.361) and
slightly helps the smaller Qwen models, so the deployable variant of the
best configuration achieves about 0.36. Second, how much of this accuracy
exceeds trivial signals? Here the honest answer is sobering: a length-only
baseline, mapping essay length through the same batch min-max onto each
prompt range, reaches macro QWK 0.523, above every LLM configuration in
this study. Note that this baseline is itself transductive in the same way
as the batch LLM variant (its min-max statistics come from the evaluation
batch), so its comparison against the frozen-calibration LLM variant is
not fully symmetric. This echoes the classic length-confound critique of AES
\cite{perelman2014}: human scores correlate strongly with length on ASAP,
and our models do not yet exploit even that signal fully. We report this
baseline prominently because it calibrates expectations for the regime we
study: sub-3B local zero-shot scorers capture a limited quality
signal, and any deployment claim that omits such baselines overstates
readiness.

\textbf{Which traits carry the signal?} Table~\ref{tab:traitcorr} ranks the
traits by the Spearman correlation of their raw scores with the resolved
human score. Content-oriented traits dominate on average: clarity and focus
(0.46), content understanding (0.44), and use of the source text (0.44)
lead the ranking, while surface conventions are weakest (0.27), consistent
with the expectation that holistic human scores weigh content more heavily
than surface conventions. The pattern is family-dependent in one telling place:
conventions correlate poorly for all three Qwen models (0.19 to 0.26) but
respectably for SmolLM2 (0.42), whose strongest trait is voice and word
choice (0.47). Trait-level reliability, not just overall accuracy, should
therefore inform model selection when rubric feedback might be shown to
students.

\begin{table}[tbp]
\centering
\caption{Spearman correlation of raw trait scores with the resolved human score, averaged over the prompts where the trait applies. Content-oriented traits carry most of the signal; surface conventions the least.}
\label{tab:traitcorr}
\begin{tabular*}{\textwidth}{@{\extracolsep{\fill}}lrrrrr}
\hline
Trait & 0.5B & 1.5B & SmolLM2 & 3B & Mean \\
\hline
clarity and focus & 0.38 & 0.50 & 0.43 & 0.52 & 0.46 \\
content understanding & 0.35 & 0.46 & 0.44 & 0.51 & 0.44 \\
use of source text & 0.34 & 0.57 & 0.39 & 0.48 & 0.44 \\
voice and word choice & 0.22 & 0.33 & 0.47 & 0.39 & 0.35 \\
ideas and content & 0.17 & 0.37 & 0.43 & 0.40 & 0.35 \\
organization & 0.21 & 0.33 & 0.42 & 0.32 & 0.32 \\
style and word choice & 0.16 & 0.28 & 0.38 & 0.40 & 0.30 \\
conventions & 0.19 & 0.26 & 0.42 & 0.21 & 0.27 \\
\hline
\end{tabular*}
\end{table}

\begin{figure}[tbp]
\centering
\includegraphics[width=\textwidth]{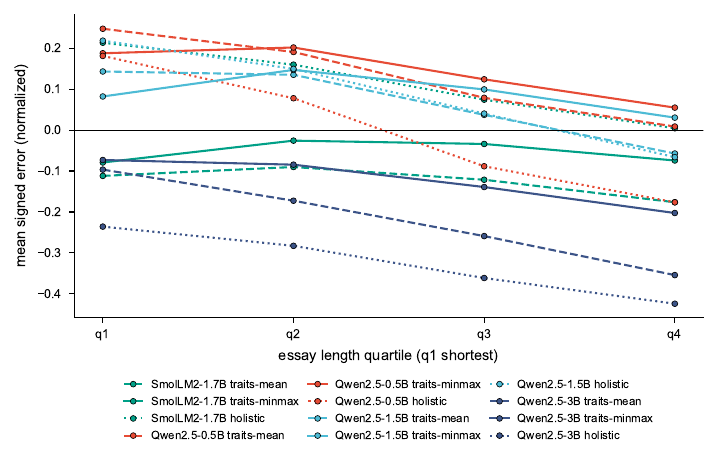}
\caption{\textbf{Length effect.} Mean signed error (normalized by prompt
score range) by essay-length quartile. Positive values overrate, negative
values underrate; eleven of twelve configurations drift downward with essay
length, and the slope is steepest under holistic prompting for three of the
four models.}
\label{fig:length}
\end{figure}

\textbf{Length bias runs opposite to LLM-judge verbosity bias.}
Fig.~\ref{fig:length} shows a systematic deployment risk: signed error
declines with essay length in eleven of twelve model-method configurations.
Under holistic prompting the mean signed error falls by 0.19 to 0.36 of the
prompt score range between the shortest and longest quartile (for example
$+0.22$ to $-0.07$ for Qwen2.5-1.5B): the two smaller Qwen models
cross from overrating to underrating, SmolLM2 declines to neutral, and
Qwen2.5-3B underrates at every quartile. This is the opposite direction to the
verbosity bias documented for large LLM judges, which favor longer
responses \cite{zheng2023}; combined with the strength of the length-only
baseline above, it says these small models \emph{under-use} the length cue
that human scores correlate with, rather than over-rewarding it. Rubric
decomposition with min-max aggregation flattens the slope dramatically for
the well-calibrated models (SmolLM2: essentially flat at $-0.08$ across
quartiles; Qwen2.5-1.5B: the drop shrinks from 0.29 to 0.05, $+0.08$ to
$+0.03$), while Qwen2.5-3B underrates across all quartiles, the same
harshness phenomenon in a second guise. In a classroom this slope has a concrete meaning: the
students who write the most would be penalized the most, and a deployment
that only checks aggregate agreement would never notice.

\textbf{Cost and error magnitude.} Trait prompting issues 3.5 calls and
about 1{,}700 prompt tokens per essay against one call and roughly 450
tokens for holistic prompting, which translates to 3.4 to 6.4 times the
wall-clock time per essay depending on the model (Table~\ref{tab:macro});
the most accurate configurations stay under four seconds per essay on the
consumer GPU, and the whole study (21{,}584 calls) consumed about 4.6
GPU-hours. In teacher-legible terms, the best configurations are off by
about 0.6 to 0.9 raw points on the narrow-range prompts 3--6 (ranges 0--3
and 0--4), by 2 to 3 points on prompt 1 (range 2--12), and by 5 to 13
points on the wide-range narrative prompts 7 and 8 (ranges up to 0--60),
where fine-grained discrimination remains far out of reach.

\section{Discussion}\label{sec:discussion}

\subsection{Practical Guidance}
For a practitioner deploying local zero-shot scoring with sub-3B FP16
open models, our results reduce to five concrete rules. \emph{(1)} Use
rubric decomposition (under batch min-max aggregation it beat holistic
prompting for every model we tested); holistic prompting wastes much of
these models' capability. \emph{(2)} Inspect the
raw trait score distribution before choosing an aggregation: a compressed,
harsh distribution calls for normalization, a well-spread one does not.
\emph{(3)} Freeze min-max statistics on a held-out calibration set rather
than the live batch; this removes the transductive dependence and, per
Table~\ref{tab:robust}, costs little accuracy, though the calibration set
must span the quality range for the mapping to be meaningful. \emph{(4)}
Monitor the shortest and longest essays explicitly: the length effect is
systematic, largest under holistic prompting for most models, and invisible
to aggregate agreement metrics. \emph{(5)} Keep the scoring formative and keep a human
in the loop: the distance to the human ceiling (0.769), and the fact that a
length-only baseline still outperforms every configuration, are not
rounding errors. We note honestly that this guidance is written for the
ed-tech engineer or researcher who builds tools for teachers; operating
this pipeline directly requires Python and GPU familiarity that most
teachers should not be expected to have.

\subsection{Limitations and Ethical Considerations}
Our evaluation covers 150 essays per prompt with essay-level and
macro-level bootstrap intervals; the resumable per-call cache makes
scaling to the full 12{,}976 essays a matter of GPU hours. ASAP essays are English,
pre-anonymized, and lack demographic metadata, so subgroup fairness (for
example toward L2 writers or dialect speakers, populations plausibly most
affected by a conventions trait) is unmeasured here and untestable on this
benchmark; Yancey et al.\ report first-language-dependent agreement for
API models \cite{yancey2023}, and the same audit is mandatory before any
real deployment. We did not probe robustness to adversarial essay content
(instruction-like text inside the essay) or strategic length manipulation,
both realistic in classrooms once students adapt to automated feedback.
The per-call cache that enables reproducibility would, with real student
writing, constitute a persistent store of student work on an unmanaged
device; deployments need explicit retention, encryption, and deletion
policies, plus student and guardian notification and a route to contest
automated feedback, in line with regulation that classifies AI systems used
to evaluate learning outcomes as high risk \cite{euaiact2024}. We deliberately exclude API models
(the research question is what schools can run themselves), restrict the
study to sub-3B models in FP16 (the same 8\,GB budget would also admit
quantized 7B models, which we leave to future work), and study
zero-shot prompting only; few-shot calibration and parameter-efficient
fine-tuning of the same small models are natural next steps, and the
formative usefulness of feedback at this accuracy level is asserted
plausibility, not something we validated with teachers or students.

\section{Conclusion}\label{sec:conclusion}
Across four sub-3B open models from two families, all run locally in
FP16, rubric-decomposed prompting is the most effective zero-shot AES
lever among the strategies we tested on consumer hardware, its benefit at
the 3B scale is fully realized only under batch min-max aggregation of
trait scores, and a deployment-safe frozen-calibration variant retains most of
that benefit. Just as
importantly, the honest anchors (a human ceiling of 0.769 and a
length-only baseline of 0.523, itself batch min-max normalized, that no
configuration beats) locate this capability precisely: real, cheap,
private, and strictly formative. The fully cached,
resumable implementation reproduces the entire study on one laptop GPU.

\bibliographystyle{splncs04o}
\bibliography{refs}

\appendix
\section*{Appendix A: Prompt Templates}
Both templates request a JSON object so that parsing is deterministic. The
system prompt for all calls is: \emph{``You are an experienced, strict but
fair teacher who grades student essays. You always answer with a single
JSON object and nothing else.''}

\medskip
\noindent\textbf{Holistic template.}
\begin{quote}\footnotesize\ttfamily
The following is [task description].\\
Read the essay and give it a single overall holistic score as an\\
integer from [lo] (worst) to [hi] (best), using the full range of\\
the scale.\\
ESSAY: """[essay text]"""\\
Respond with only a JSON object of the form\\
\{"score": <integer between [lo] and [hi]>\}.
\end{quote}

\medskip
\noindent\textbf{Trait template} (one call per trait).
\begin{quote}\footnotesize\ttfamily
The following is [task description].\\
Evaluate ONLY the trait "[trait name]": [trait description].\\
Give an integer score from 0 (very poor) to 10 (excellent) for\\
this trait alone, ignoring all other aspects of the essay.\\
ESSAY: """[essay text]"""\\
Respond with only a JSON object of the form\\
\{"score": <integer between 0 and 10>\}.
\end{quote}

\end{document}